\def\year{2020}\relax
\documentclass[letterpaper]{article} 
\usepackage{aaai20}  
\usepackage{times}  
\usepackage{helvet} 
\usepackage{courier}  
\usepackage[hyphens]{url}  
\usepackage{graphicx} 
\def\UrlFont{\rm}  
\usepackage{graphicx}  
\usepackage{booktabs}
\usepackage{xcolor}
\usepackage{xspace}

\newcommand{\academy}{\emph{Football Academy}\xspace}
\newcommand{\engine}{\emph{Football Engine}\xspace}
\newcommand{\benchmarks}{\emph{Football Benchmarks}\xspace}

\usepackage{subcaption}
\usepackage{listings}
\usepackage{natbib}

\title{Google Research Football: A Novel Reinforcement Learning Environment}
\author{\large Karol Kurach\thanks{Indicates equal authorship. Correspondence to Karol Kurach (kkurach@google.com).}\hspace{0.2cm} Anton Raichuk$^\star$ \hspace{0.2cm} Piotr Sta{\accent19 n}czyk$^\star$ \hspace{0.2cm} Micha\l{} Zaj{\accent24 a}c\thanks{Student at Jagiellonian University, work done during internship at Google Brain.}\\ \\%
\large\textbf{Olivier Bachem\hspace{0.2cm} Lasse Espeholt\hspace{0.2cm} Carlos Riquelme\hspace{0.2cm} Damien Vincent} \\ \\%
\large\textbf{Marcin Michalski\hspace{0.2cm} Olivier Bousquet\hspace{0.2cm} Sylvain Gelly} \\ \\
Google Research, Brain Team 
}

\begin{document}

\maketitle

\begin{abstract}
\looseness=-1 Recent progress in the field of reinforcement learning has been accelerated by virtual learning environments such as video games, where novel algorithms and ideas can be quickly tested in a safe and reproducible manner. We introduce the \emph{Google Research Football Environment}, a new reinforcement learning environment where agents are trained to play football in an advanced, physics-based 3D simulator.
The resulting environment is challenging, easy to use and customize,
and it is available under a permissive open-source license. In addition, it provides support for multiplayer and multi-agent experiments. We propose three full-game scenarios of varying difficulty with the \emph{Football Benchmarks} and report baseline results for three commonly used reinforcement algorithms (IMPALA, PPO, and Ape-X DQN).
We also provide a diverse set of simpler scenarios with the \emph{Football Academy} and showcase several promising research directions. 
\end{abstract}
\section{Introduction}

The goal of reinforcement learning (RL) is to train smart agents that can interact with their environment and solve complex tasks \citep{sutton2018reinforcement}. Real-world applications include robotics \citep{haarnoja2018soft}, self-driving cars \citep{bansal2018chauffeurnet}, and control problems such as increasing the power efficiency of data centers  \citep{lazic2018data}. 
Yet, the rapid progress in this field has been fueled by making agents play games such as the iconic Atari console games \citep{bellemare2013arcade,mnih2013playing}, the ancient game of Go \citep{silver2016mastering}, or professionally played video games like Dota 2 \citep{openai_dota} or Starcraft II \citep{vinyals2017starcraft}.
The reason for this is simple: games provide challenging environments where new algorithms and ideas can be quickly tested in a safe and reproducible manner.

\begin{figure}[t]
    \centering
    \includegraphics[width=\columnwidth,trim={2px 2px 2px 35px},clip]{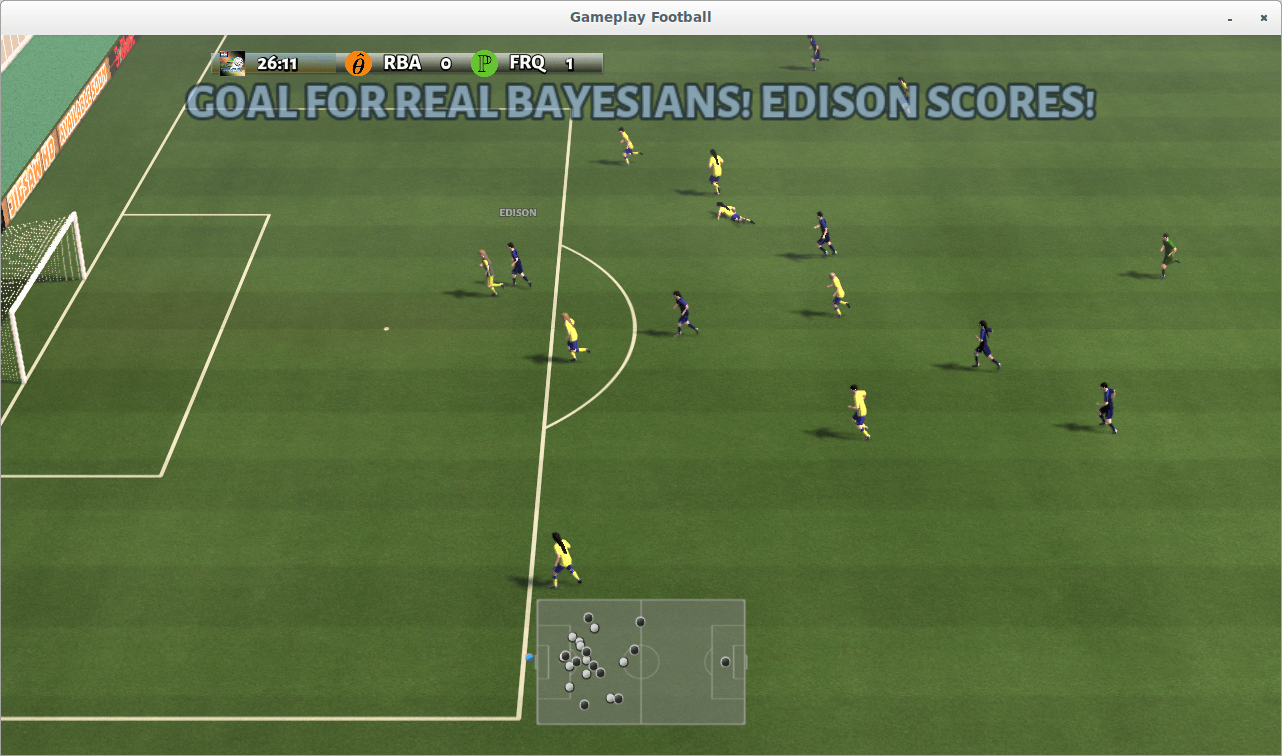}
    \caption{The \emph{Google Research Football Environment} (\texttt{github.com/google-research/football}) provides a novel reinforcement learning environment where agents are trained to play football in an advance, physics based 3D simulation.}
    \label{fig:main}
\end{figure}

While a variety of reinforcement learning environments exist, they often come with a few drawbacks for research, which we discuss in detail in the next section.
For example, they may either be too easy to solve for state-of-the-art algorithms or require access to large amounts of computational resources.
At the same time, they may either be (near-)deterministic or there may even be a known model of the environment (such as in Go or Chess).
Similarly, many learning environments are inherently single player by only modeling the interaction of an agent with a fixed environment or they focus on a single aspect of reinforcement learning such as continuous control or safety.
Finally, learning environments may have restrictive licenses or depend on closed source binaries.

This highlights the need for a RL environment that is not only challenging from a learning standpoint and customizable in terms of difficulty but also accessible for research both in terms of licensing and in terms of required computational resources. 
Moreover, such an environment should ideally provide the tools to a variety of current reinforcement learning research topics such as the impact of stochasticity, self-play, multi-agent setups and model-based reinforcement learning, while also requiring smart decisions, tactics, and strategies at multiple levels of abstraction. 

\begin{figure*}[t]
    \begin{subfigure}{.32\textwidth}
        \centering
        \includegraphics[width=\columnwidth,trim={2px 2px 2px 35px},clip]{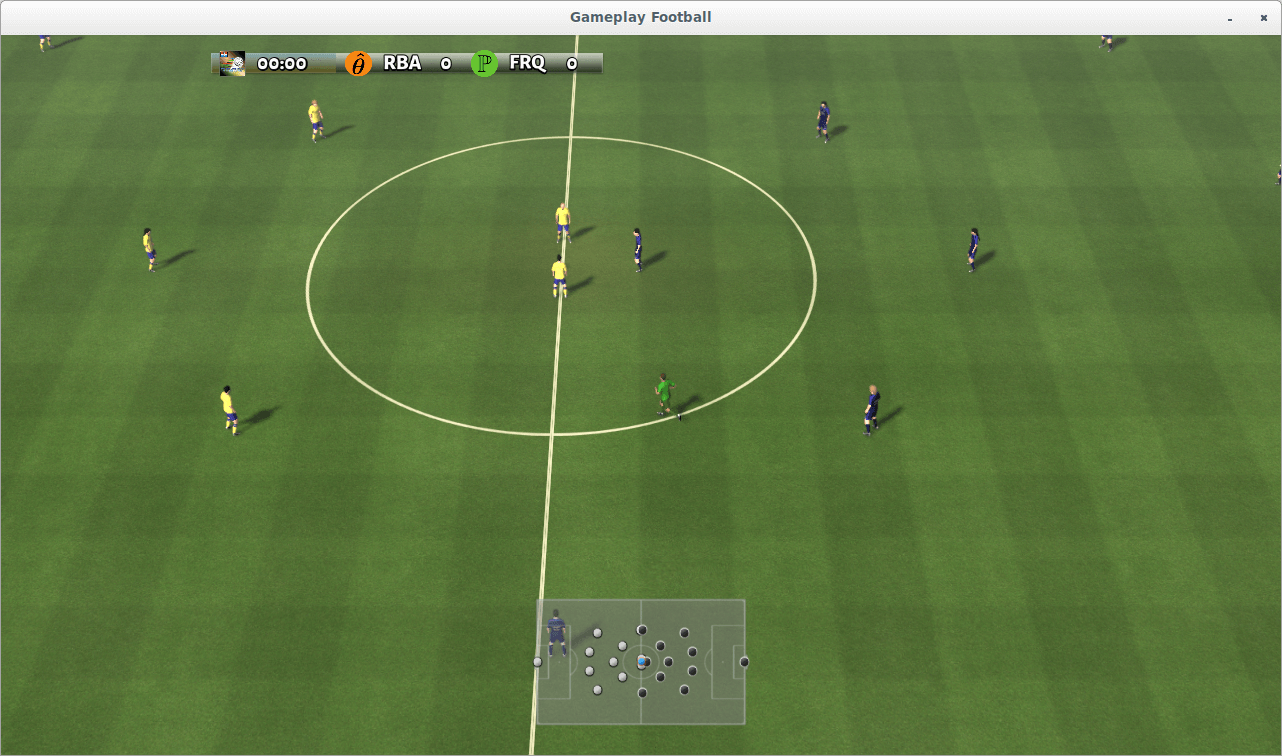}
        \caption{Kickoff}
    \end{subfigure}\hfill
    \begin{subfigure}{.32\textwidth}
        \centering
        \includegraphics[width=\columnwidth,trim={2px 2px 2px 35px},clip]{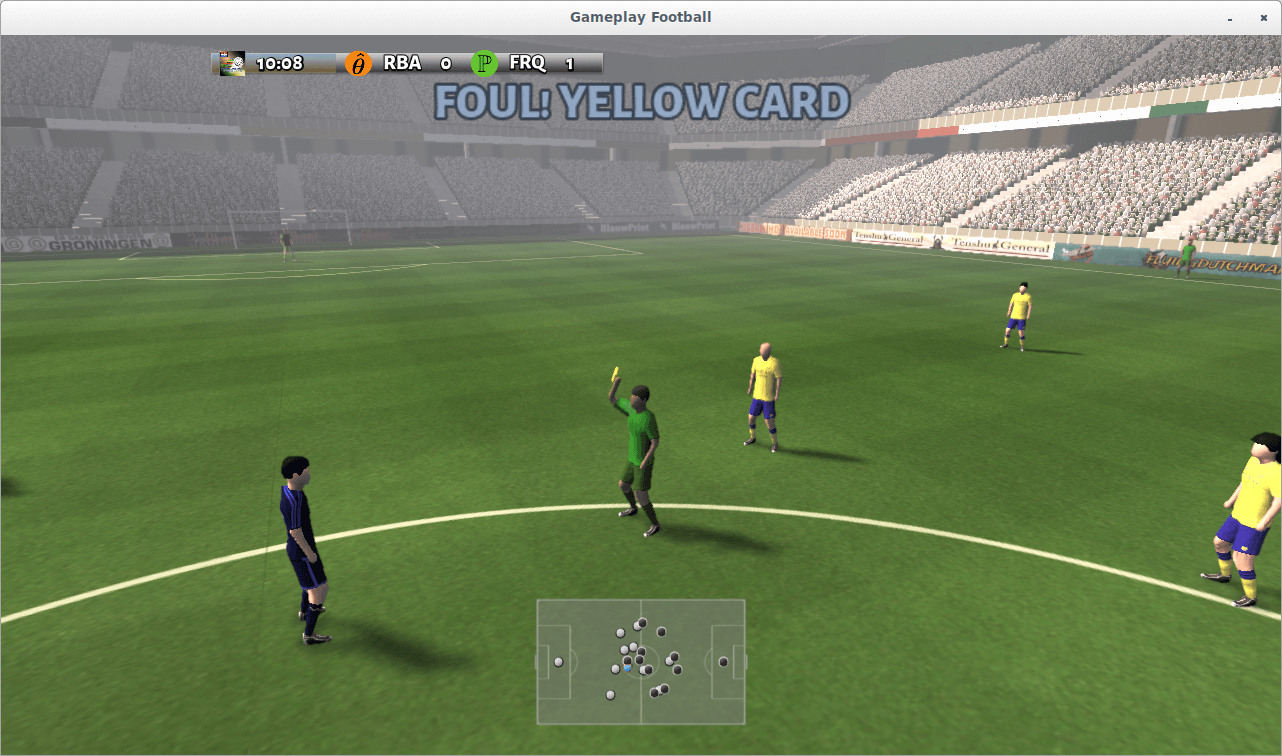}
        \caption{Yellow card}
    \end{subfigure}\hfill
    \begin{subfigure}{.32\textwidth}
        \centering
        \includegraphics[width=\columnwidth,trim={2px 2px 2px 35px},clip]{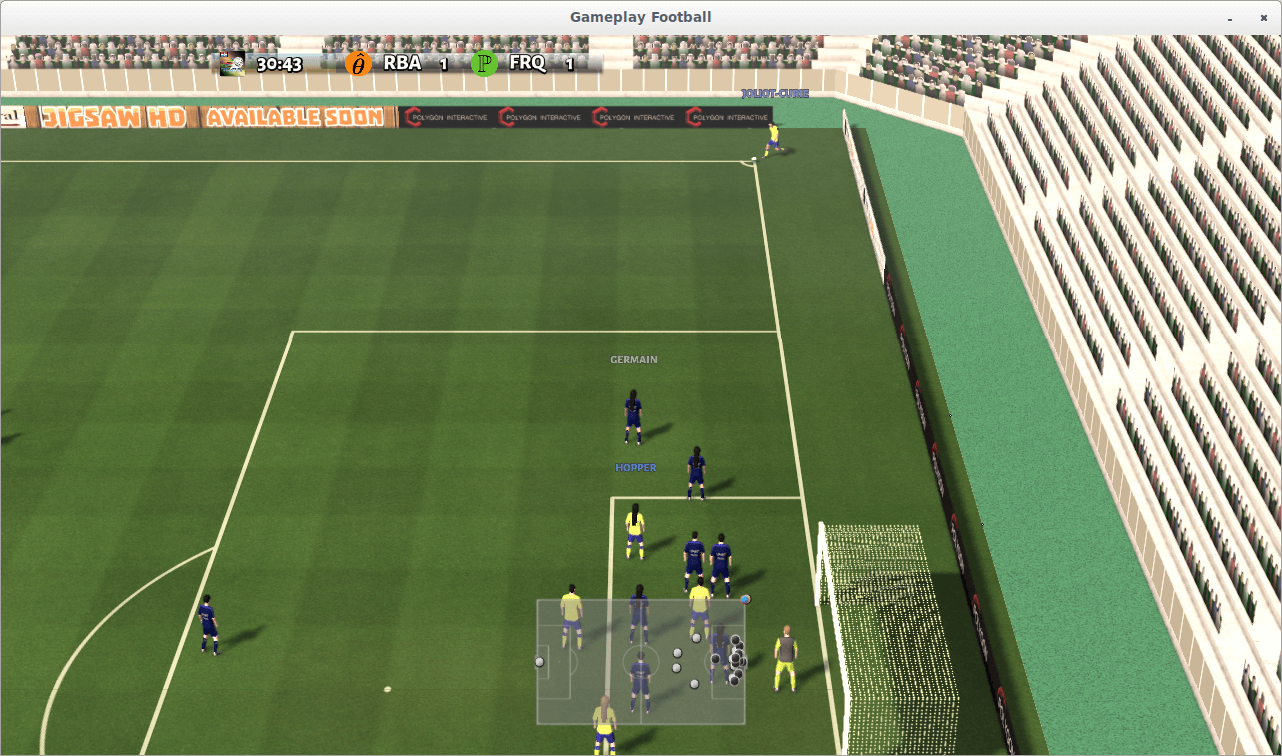}
        \caption{Corner kick}
    \end{subfigure}
    \caption{The \engine is an advanced football simulator that supports all the major football rules such as (a) kickoffs (b) goals, fouls, cards, (c) corner kicks, penalty kicks, and offside.}
    \label{fig:game_features}
\end{figure*}

\paragraph{Contributions} 
In this paper, we propose the \emph{Google Research Football Environment}, a novel open-source reinforcement learning environment where agents learn to play one of the world's most popular sports: football (a.k.a.\ soccer). 
Modeled after popular football video games, the Football Environment provides a physics-based 3D football simulation where agents have to control their players, learn how to pass in between them and how to overcome their opponent's defense in order to score goals. 
This provides a challenging RL problem as football requires a natural balance between short-term control, learned concepts such as passing, and high level strategy.
As our key contributions, we
\begin{itemize}
\item provide the \engine, a highly-optimized game engine that simulates the game of football,
\item propose the \benchmarks, a versatile set of benchmark tasks of varying difficulties that can be used to compare different algorithms,
\item propose the \academy, a set of progressively harder and diverse reinforcement learning scenarios,
\item evaluate state-of-the-art algorithms on both the \benchmarks and the \academy, providing an extensive set of reference results for future comparison, 
\item provide a simple API to completely customize and define new football reinforcement learning scenarios, and
\item showcase several promising research directions in this environment, \emph{e.g.} the multi-player and multi-agent settings.
\end{itemize}
\section{Motivation and Other Related Work}
There are a variety of reinforcement learning environments that have accelerated research in recent years.
However, existing environments exhibit a variety of drawbacks that we address with the \emph{Google Research Football Environment}:

\looseness-1\paragraph{Easy to solve.}
With the recent progress in RL, many commonly used scenarios can now be solved to a reasonable degree in just a few hours with well-established algorithms.
For instance, ${\sim} 50$ commonly used Atari games in the \emph{Arcade Learning Environment} \citep{bellemare2013arcade} are routinely solved to super-human level \citep{hessel2018rainbow}.
The same applies to the \emph{DeepMind Lab} \citep{beattie2016deepmind}, a navigation-focused maze environment that provides a number of relatively simple tasks with a first person viewpoint.

\paragraph{Computationally expensive.}
On the other hand, training agents in recent video-game simulators often requires substantial computational resources that may not be available to a large fraction of researchers due to combining hard games, long episodes, and high-dimensional inputs (either in the form of pixels, or hand-crafted representations).
For example, the \emph{StarCraft II Learning Environment} \citep{vinyals2017starcraft} provides an API to \emph{Starcraft II}, a well-known real-time strategy video game, as well as to a few mini-games which are centered around specific tasks in the game.

\paragraph{Lack of stochasticity.}
\looseness=-2 The real-world is not deterministic which motivates the need to develop algorithms that can cope with and learn from stochastic environments.
Robots, self-driving cars, or data-centers require robust policies that account for uncertain dynamics.
Yet, some of the most popular simulated environments -- like the Arcade Learning Environment -- are deterministic. 
While techniques have been developed to add artificial randomness to the environment (like skipping a random number of initial frames or using sticky actions), this randomness may still be too structured and easy to predict and incorporate during training \citep{machado2018revisiting,hausknecht2015impact}.
It remains an open question whether modern RL approaches such as self-imitation generalize from the deterministic setting to stochastic environments \citep{guo2018generative}.

\paragraph{Lack of open-source license.}
Some advanced physics simulators offer licenses that may be subjected to restrictive use terms \citep{todorov2012mujoco}.
Also, some environments such as StarCraft require access to a closed-source binary.
In contrast, open-source licenses enable researchers to inspect the underlying game code and to modify environments if required to test new research ideas. 

\paragraph{Known model of the environment.}
Reinforcement learning algorithms have been successfully applied to board games such as Backgammon \citep{tesauro1995temporal}, Chess \citep{hsu2004behind}, or Go \citep{silver2016mastering}.
Yet, current state-of-the-art algorithms often exploit that the rules of these games (\emph{i.e.}, the model of the environment) are specific, known and can be encoded into the approach.
As such, this may make it hard to investigate learning algorithms that should work in environments that can only be explored through interactions.

\paragraph{Single-player.}
In many available environments such as Atari, one only controls a single agent. 
However, some modern real-world applications involve a number of agents under either centralized or distributed control. The different agents can either collaborate or compete, creating additional challenges. 
A well-studied special case is an agent competing against another agent in a zero sum game.
In this setting, the opponent can adapt its own strategy, and the agent has to be robust against a variety of opponents. 
Cooperative multi-agent learning also offers many opportunities and challenges, such as communication between agents, agent behavior specialization, or robustness to the failure of some of the agents.
Multiplayer environments with collaborative or competing agents can help foster research around those challenges.

\paragraph{Other football environments.}
There are other available football simulators, such as the \emph{RoboCup Soccer Simulator} \citep{kitano1995robocup,kitano1997robocup}, and the \emph{DeepMind MuJoCo Multi-Agent Soccer Environment} \citep{liu2019emergent}.
In contrast to these environments, the \emph{Google Research Football Environment} focuses on high-level actions instead of low-level control of a physics simulation of robots (such as in the RoboCup Simulation 3D League). 
Furthermore, it provides many useful settings for reinforcement learning, e.g. the single-agent and multi-agent settings as well as single-player and multiplayer player modes. 
\emph{Google Research Football} also provides ways to adjust difficulty, both via a strength-adjustable opponent and via diverse and customizable scenarios in Football Academy, and provides several specific features for reinforcement learning research, e.g., OpenAI gym compatibility, different rewards, different representations, and the option to turn on and off stochasticity.

\paragraph{Other related work.}
Designing rich learning scenarios is challenging, and resulting environments often provide a useful playground for research questions centered around a specific reinforcement learning set of topics.
For instance, the \emph{DeepMind Control Suite} \citep{tassa2018deepmind} focuses on continuous control,
the \emph{AI Safety Gridworlds} \citep{leike2017ai} on learning safely, whereas the \emph{Hanabi Learning Environment} \citep{bard2019hanabi} proposes a multi-agent setup.
As a consequence, each of these environments are better suited for testing algorithmic ideas involving a limited but well-defined set of research areas.

\section{Football Engine}
The Football Environment is based on the Football Engine, an advanced football simulator built around a heavily customized version of the publicly available \emph{GameplayFootball} simulator \citep{gameplayfootball}.
The engine simulates a complete football game, and includes the most common football aspects, such as goals, fouls, corners, penalty kicks, or off-sides (see Figure~\ref{fig:game_features} for a few examples).

\paragraph{Supported Football Rules.}
The engine implements a full football game under standard rules, with 11 players on each team.
These include goal kicks, side kicks, corner kicks, both yellow and red cards, offsides, handballs and penalty kicks.
The length of the game is measured in terms of the number of frames, and the default duration of a full game is $3000$ ($10$ frames per second for $5$ minutes).
The length of the game, initial number and position of players can also be edited in customized scenarios (see \academy below).
Players on a team have different statistics\footnote{Although players differ within a team, both teams have exactly the same set of players, to ensure a fair game.}, such as speed or accuracy and get tired over time.

\paragraph{Opponent AI Built-in Bots.}
The environment controls the opponent team by means of a rule-based bot, which was provided by the original \emph{GameplayFootball} simulator \citep{gameplayfootball}.
The difficulty level $\theta$ can be smoothly parameterized between 0 and 1, by speeding up or slowing down the bot reaction time and decision making.
Some suggested difficulty levels correspond to: easy ($\theta = 0.05$), medium ($\theta = 0.6$), and hard ($\theta = 0.95$).
For self-play, one can replace the opponent bot with any trained model.

Moreover, by default, our non-active players are also controlled by another rule-based bot.
In this case, the behavior is simple and corresponds to reasonable football actions and strategies, such as running towards the ball when we are not in possession, or move forward together with our active player.
In particular, this type of behavior can be turned off for future research on cooperative multi-agents if desired.

\paragraph{State \& Observations.}
We define as \emph{state} the complete set of data that is returned by the environment after actions are performed.
On the other hand, we define as \emph{observation} or \emph{representation} any transformation of the state that is provided as input to the control algorithms.
The definition of the state contains information such as the ball position and possession, coordinates of all players, the active player, the game state (tiredness levels of players, yellow cards, score, etc) and the current pixel frame.

\looseness=-1We propose three different representations.
Two of them (pixels and SMM) can be \emph{stacked} across multiple consecutive time-steps (for instance, to determine the ball direction), or unstacked, that is, corresponding to the current time-step only.
Researchers can easily define their own representations based on the environment state by creating wrappers similar to the ones used for the observations below.

\textit{Pixels.} The representation consists of a $1280 \times 720$ RGB image corresponding to the rendered screen.
This includes both the scoreboard and a small map in the bottom middle part of the frame from which the position of all players can be inferred in principle.

\textit{Super Mini Map.}\label{sec:smm} The SMM representation consists of four $72 \times 96$ matrices encoding information about the home team, the away team, the ball, and the active player respectively. The encoding is binary, representing whether there is a player or ball in the corresponding coordinate.

\looseness=-1\textit{Floats.} The floats representation provides a compact encoding and consists of a 115-dimensional vector summarizing many aspects of the game, such as players coordinates, ball possession and direction, active player, or game mode.

\paragraph{Actions.}
The actions available to an individual agent (player) are displayed in Table~\ref{tab:actions}.
They include standard move actions (in $8$ directions), and different ways to kick the ball (short and long passes, shooting, and high passes that can't be easily intercepted along the way).
Also, players can sprint (which affects their level of tiredness), try to intercept the ball with a slide tackle or dribble if they posses the ball. 
We experimented with an action to switch the active player in defense (otherwise, the player with the ball must be active).
However, we observed that policies tended to exploit this action to return control to built-in AI behaviors for non-active players, and we decided to remove it from the action set.
We do \emph{not} implement randomized sticky actions.
Instead, once executed, moving and sprinting actions are sticky and continue until an explicit stop action is performed (Stop-Moving and Stop-Sprint respectively).

\paragraph{Rewards.}
The \engine includes two reward functions that can be used out-of-the-box: \textsc{Scoring} and \textsc{Checkpoint}.
It also allows researchers to add custom reward functions using wrappers which can be used to investigate reward shaping approaches.

\textsc{Scoring} corresponds to the natural reward where each team obtains a $+1$ reward when scoring a goal, and a $-1$ reward when conceding one to the opposing team.
The \textsc{Scoring} reward can be hard to observe during the initial stages of training, as it may require a long sequence of consecutive events: overcoming the defense of a potentially strong opponent, and scoring against a keeper.

\textsc{Checkpoint} is a (shaped) reward that specifically addresses the sparsity of \textsc{Scoring} by encoding the domain knowledge that scoring is aided by advancing across the pitch:
It augments the \textsc{Scoring} reward with an additional auxiliary reward contribution for moving the ball close to the opponent's goal in a controlled fashion.
More specifically, we divide the opponent's field in 10 checkpoint regions according to the Euclidean distance to the opponent goal.
Then, the first time the agent's team possesses the ball in each of the checkpoint regions, the agent obtains an additional reward of $+0.1$.
This extra reward can be up to $+1$, \emph{i.e.}, the same as scoring a single goal.
Any non-collected checkpoint reward is also added when scoring in order to avoid penalizing agents that do not go through all the checkpoints before scoring (\emph{i.e.}, by shooting from outside a checkpoint region).
Finally, checkpoint rewards are only given once per episode.

\paragraph{Accessibility.}
Researchers can directly inspect the game by playing against each other or by dueling their agents.
The game can be controlled by means of both keyboards and gamepads.
Moreover, replays of several rendering qualities can be automatically stored while training, so that it is easy to inspect the policies agents are learning.

\begin{figure}[t]
    \centering
    \includegraphics[width=0.9\columnwidth]{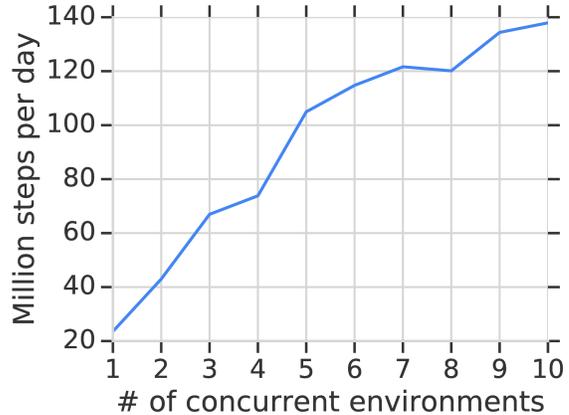}
    \caption{Number of steps per day versus number of concurrent environments for the \engine on a hexa-core Intel Xeon W-2135 CPU with 3.70GHz.}
    \label{fig:speed}
\end{figure}

\paragraph{Stochasticity.}
In order to investigate the impact of randomness, and to simplify the tasks when desired, the environment can run in either  stochastic or deterministic mode.
The former, which is enabled by default, introduces several types of randomness: for instance, the same shot from the top of the box may lead to a different number of outcomes.
In the latter, playing a fixed policy against a fixed opponent always results in the same sequence of actions and states.

\paragraph{API \& Sample Usage.}
The Football Engine is out of the box compatible with the widely used OpenAI Gym API \citep{brockman2016openai}.
Below we show example code that runs a random agent on our environment.

\begin{lstlisting}
import gfootball.env as football_env

env = football_env.create_environment(
    env_name='11_vs_11_stochastic', 
    render=True)
env.reset()
done = False
while not done:
    action = env.action_space.sample()
    observation, reward, done, info = \
        env.step(action)
\end{lstlisting}

\paragraph{Technical Implementation \& Performance.}
The Football Engine is written in highly optimized C++ code, allowing it to be run on commodity machines both with GPU and without GPU-based rendering enabled. This allows it to obtain a performance of approximately $140$ million steps per day on a single hexacore machine (see Figure~\ref{fig:speed}).

\begin{table}[b]
\centering
\caption{Action Set}
\label{tab:actions}
{
\scriptsize
\begin{tabular}{cccc}
\toprule
      Top & Bottom & Left & Right \\
      Top-Left & Top-Right & Bottom-Left & Bottom-Right \\
      Short Pass & High Pass & Long Pass & Shot \\
      Do-Nothing & Sliding & Dribble & Stop-Dribble \\
      Sprint &
      Stop-Moving & 
      Stop-Sprint &
      --- \\
\bottomrule
\end{tabular}
}
\end{table}

\begin{figure*}[t]
    \centering
    \includegraphics[width=\textwidth]{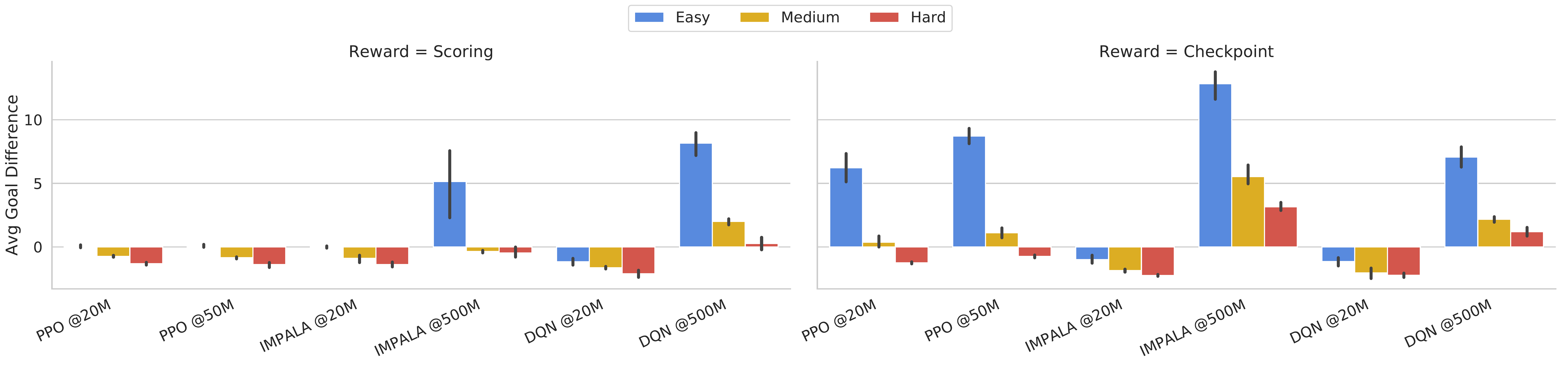}
    \caption{Average Goal Difference on the \benchmarks for IMPALA, PPO and Ape-X DQN with both the \textsc{Scoring} and \textsc{Checkpoint} rewards. Error bars represent $95$\% bootstrapped confidence intervals. Results for GRF v$2$.x}
    \label{fig:challenges_both_rewards}
\end{figure*}

\section{Football Benchmarks}
The \engine is an efficient, flexible and highly customizable learning environment with many features that lets researchers try a broad range of new ideas.
To facilitate fair comparisons of different algorithms and approaches in this environment, we also provide a set of pre-defined benchmark tasks that we call the \benchmarks. 
Similar to the Atari games in the \emph{Arcade Learning Environment}, in these tasks, the agent has to interact with a fixed environment and maximize its episodic reward by sequentially choosing suitable actions based on observations of the environment.

The goal in the \benchmarks is to win a full game\footnote{We define an 11 versus 11 full game to correspond to 3000 steps in the environment, which amounts to 300 seconds if rendered at a speed of 10 frames per second.} against the opponent bot provided by the engine.
We provide three versions of the \benchmarks that only differ in the strength of the opponent AI as described in the last section: the easy, medium, and hard benchmarks.
This allows researcher to test a wide range of research ideas under different computational constraints such as single machine setups or powerful distributed settings. 
We expect that these benchmark tasks will be useful for investigating current scientific challenges in reinforcement learning such as sample-efficiency, sparse rewards, or model-based approaches.

\subsection{Experimental Setup}
As a reference, we provide benchmark results for three state-of-the-art reinforcement learning algorithms: PPO \citep{schulman2017proximal} and IMPALA \citep{espeholt2018impala} which are popular policy gradient methods, and Ape-X DQN~\citep{Horgan2018DistributedPE}, which is a modern DQN implementation. 
We run PPO in multiple processes on a single machine, while IMPALA and DQN are run on a distributed cluster with $500$ and $150$ actors respectively.

In all benchmark experiments, we use the stacked Super Mini Map representation \ref{sec:smm} and the same network architecture. 
We consider both the \textsc{Scoring} and \textsc{Checkpoint} rewards.
The tuning of hyper-parameters is done using easy scenario, and we follow the same protocol for all algorithms to ensure fairness of comparison.
After tuning, for each of the six considered settings (three \benchmarks and two reward functions), we run five random seeds and average the results.
For the technical details of the training setup and the used architecture and hyperparameters, we refer to the Appendix.

\begin{figure*}[t!]
    \begin{subfigure}{.32\textwidth}
        \centering
        \includegraphics[width=\linewidth,trim={2px 2px 2px 35px},clip]{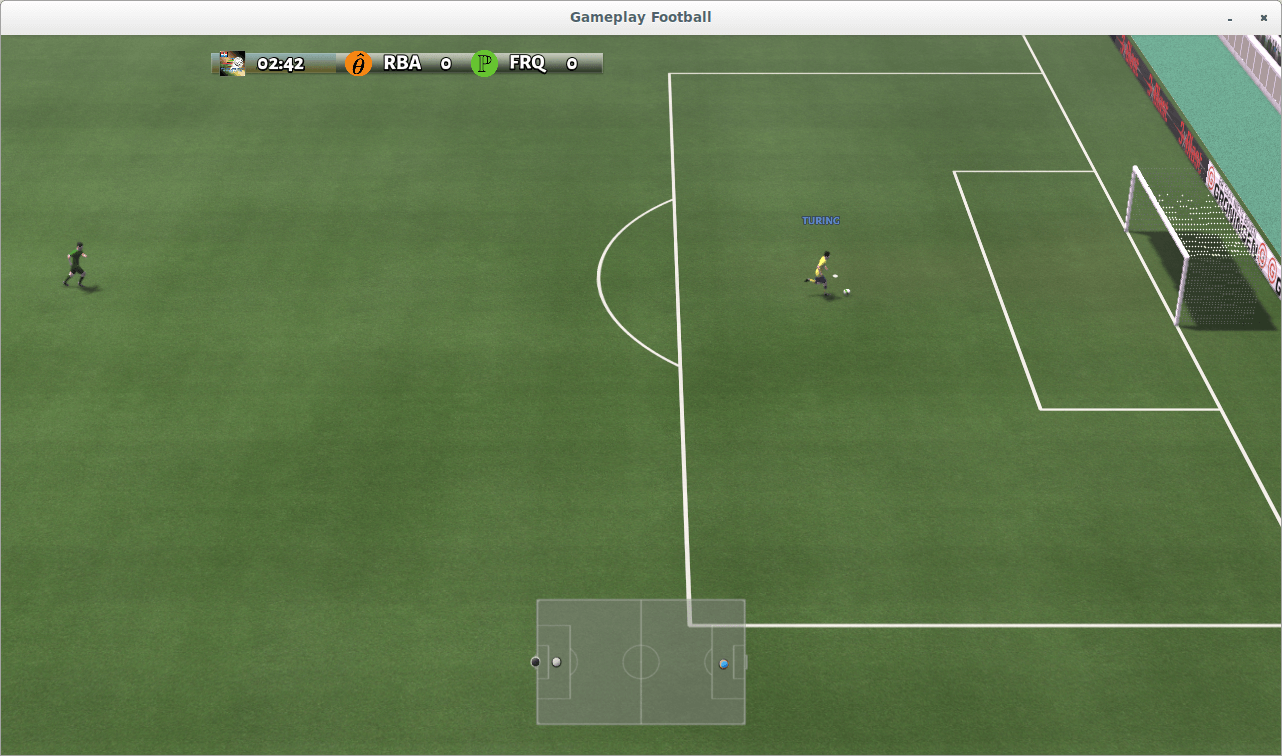}
        \caption{Empty Goal Close}
    \end{subfigure}\hfill
    \begin{subfigure}{.32\textwidth}
        \centering
        \includegraphics[width=\linewidth,trim={2px 2px 2px 35px},clip]{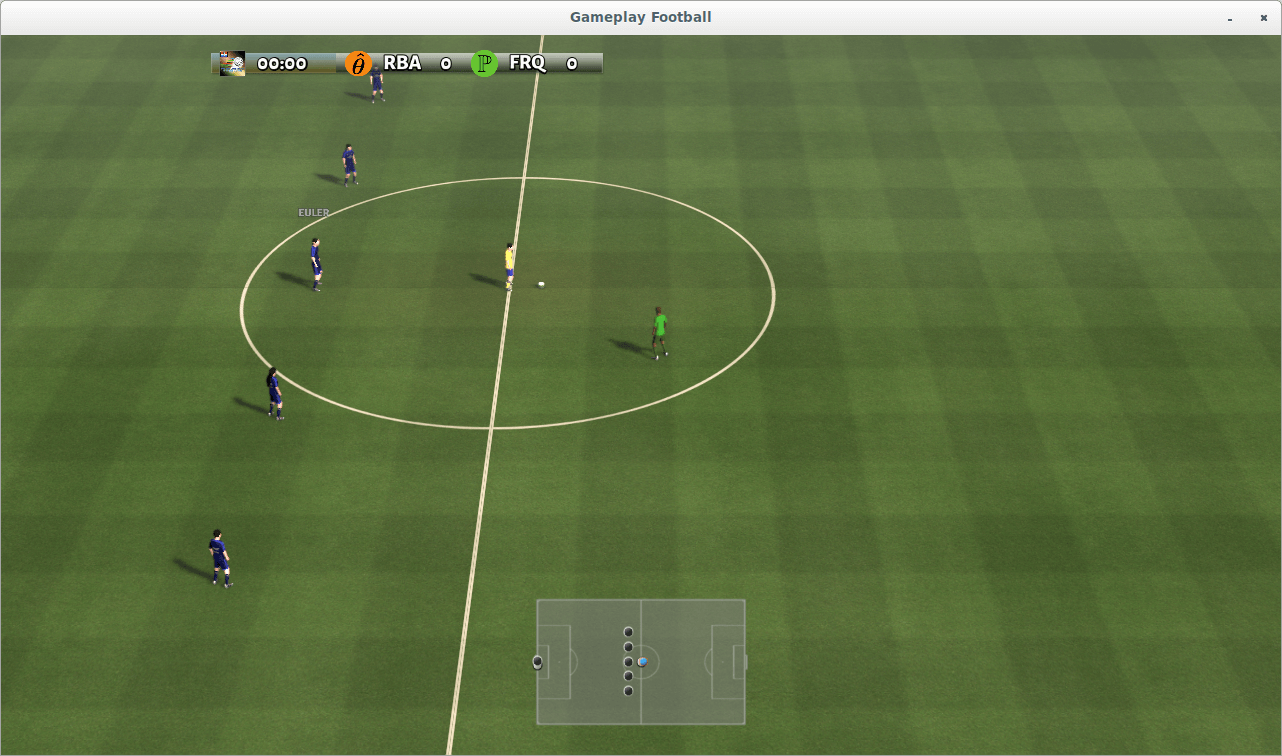}
        \caption{Run to Score}
    \end{subfigure}\hfill
    \begin{subfigure}{.32\textwidth}
        \centering
        \includegraphics[width=\linewidth,trim={2px 2px 2px 35px},clip]{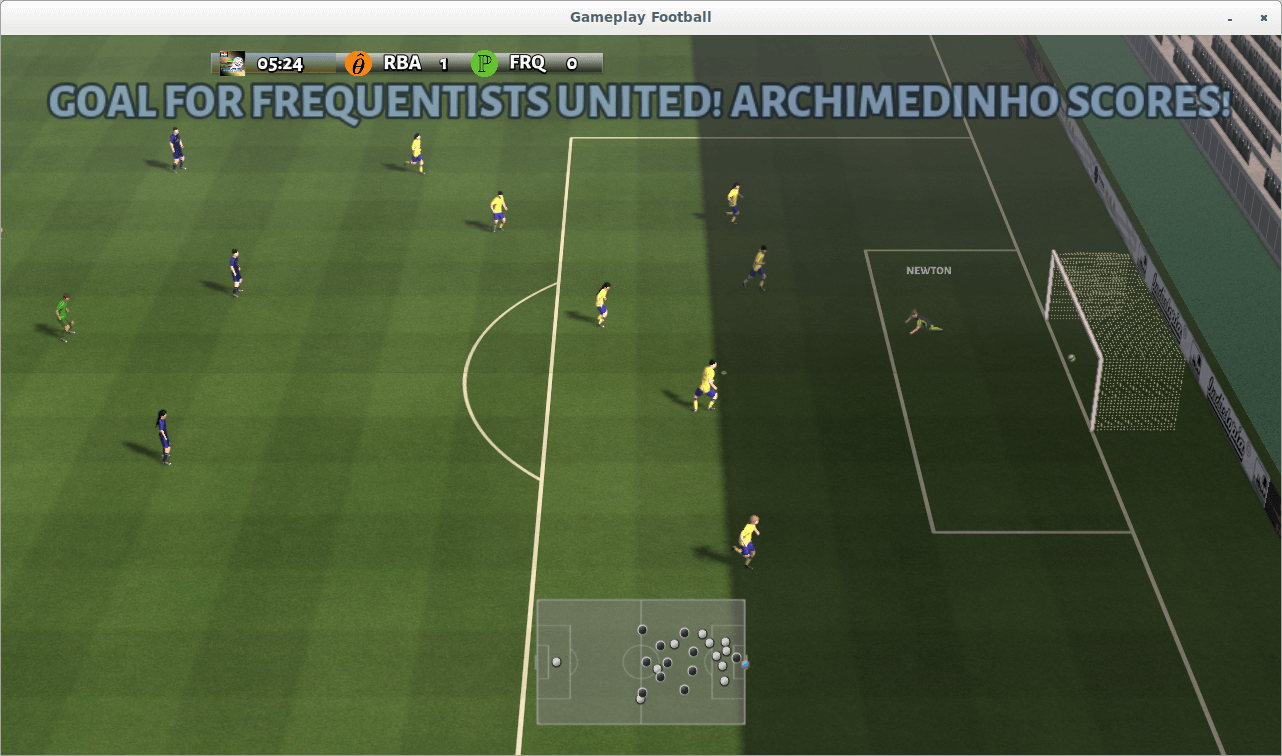}
        \caption{11 vs 11 with Lazy Opponents}
    \end{subfigure}\hfill
    \begin{subfigure}{.32\textwidth}
        \centering
        \includegraphics[width=\linewidth,trim={2px 2px 2px 35px},clip]{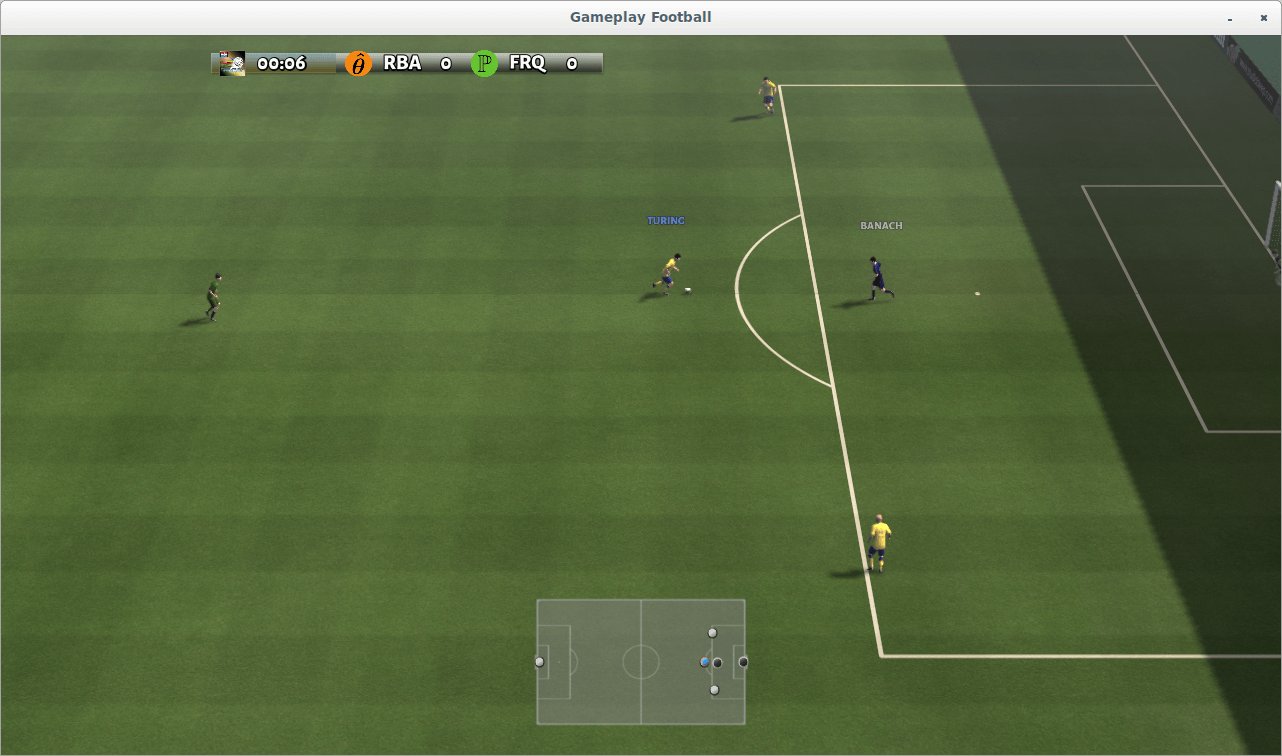}
        \caption{3 vs 1 with Keeper}
    \end{subfigure}\hfill
    \begin{subfigure}{.32\textwidth}
        \centering
        \includegraphics[width=\linewidth,trim={2px 2px 2px 35px},clip]{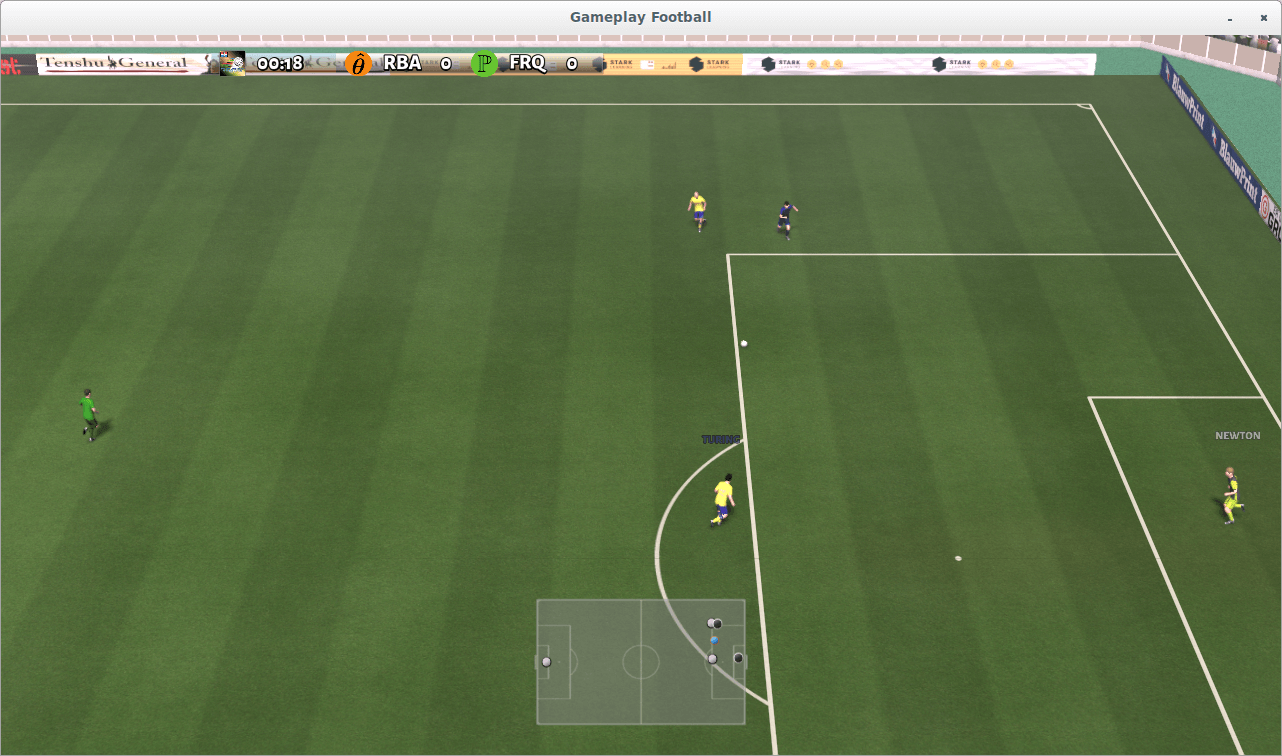}
        \caption{Pass and Shoot}
    \end{subfigure}\hfill
    \begin{subfigure}{.32\textwidth}
        \centering
        \includegraphics[width=\linewidth,trim={2px 2px 2px 35px},clip]{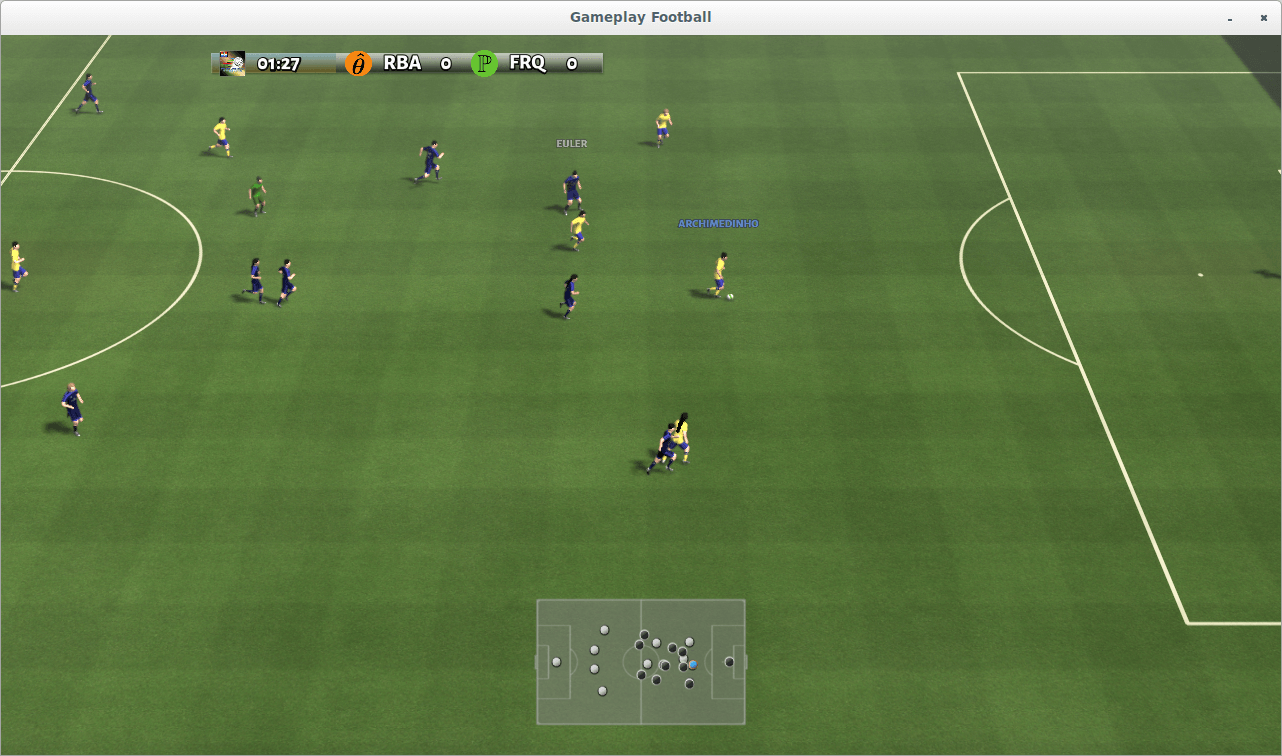}
        \caption{Easy Counter-attack}
    \end{subfigure}
    \caption{Example of \academy scenarios.}
    \label{fig:academy_scenarios}
\end{figure*}

\subsection{Results}
\looseness=-1The experimental results\footnote{All results in this paper are for the versions v$2$.x of the GRF.} for the \benchmarks are shown in Figure~\ref{fig:challenges_both_rewards}. 
It can be seen that the environment difficulty significantly affects the training complexity and the average goal difference. 
The medium benchmark can be beaten by DQN and IMPALA with 500M training steps (albeit only barely with the \textsc{Scoring} reward).
The hard benchmark is even harder, and requires the \textsc{Checkpoint} reward and 500M training steps for achieving a positive score.
We observe that the \textsc{Checkpoint} reward function appears to be very helpful for speeding up the training for policy gradient methods but does not seem to benefit as much the Ape-X DQN as the performance is similar with both the \textsc{Checkpoint} and \textsc{Scoring} rewards.
We conclude that the \benchmarks provide interesting reference problems for research and that there remains a large headroom for progress, in particular in terms of performance and sample efficiency on the harder benchmarks.
\begin{figure*}[t]
    \centering
    \includegraphics[scale=0.24]{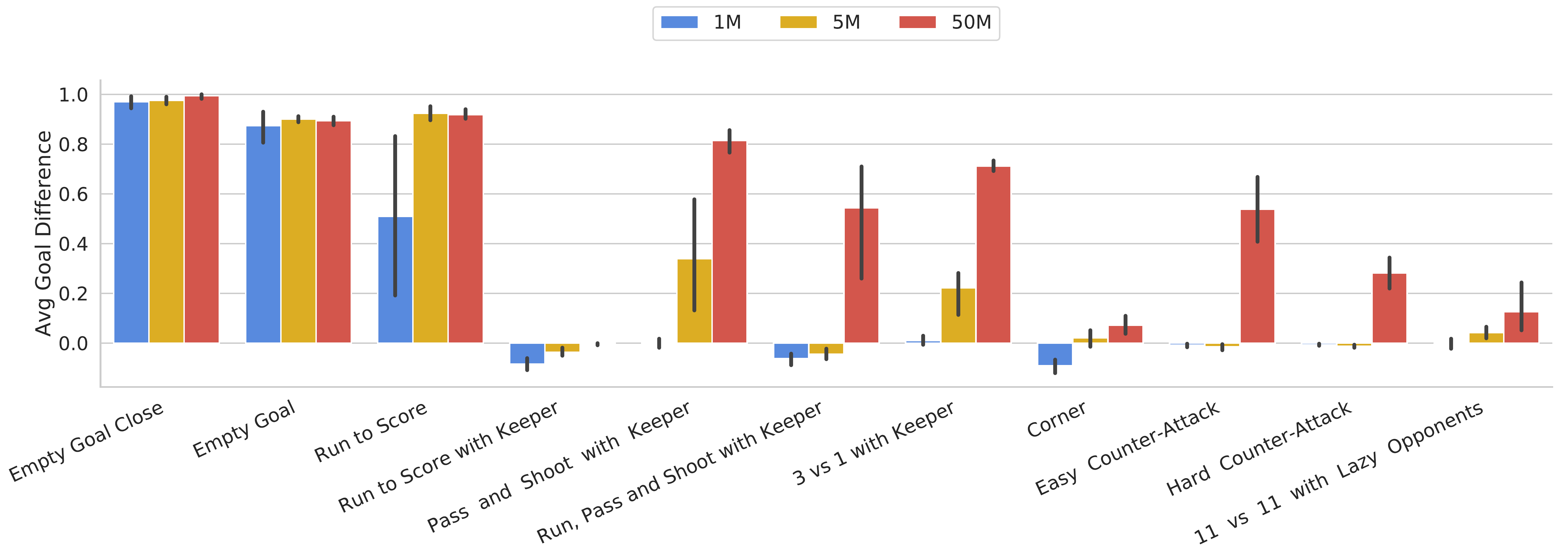}
    \caption{Average Goal Difference on  \academy for IMPALA with \textsc{Scoring} reward.}
    \label{fig:academy_impala_scoring}
    
    \vspace{0.23cm}
    
    \centering
    \includegraphics[scale=0.24]{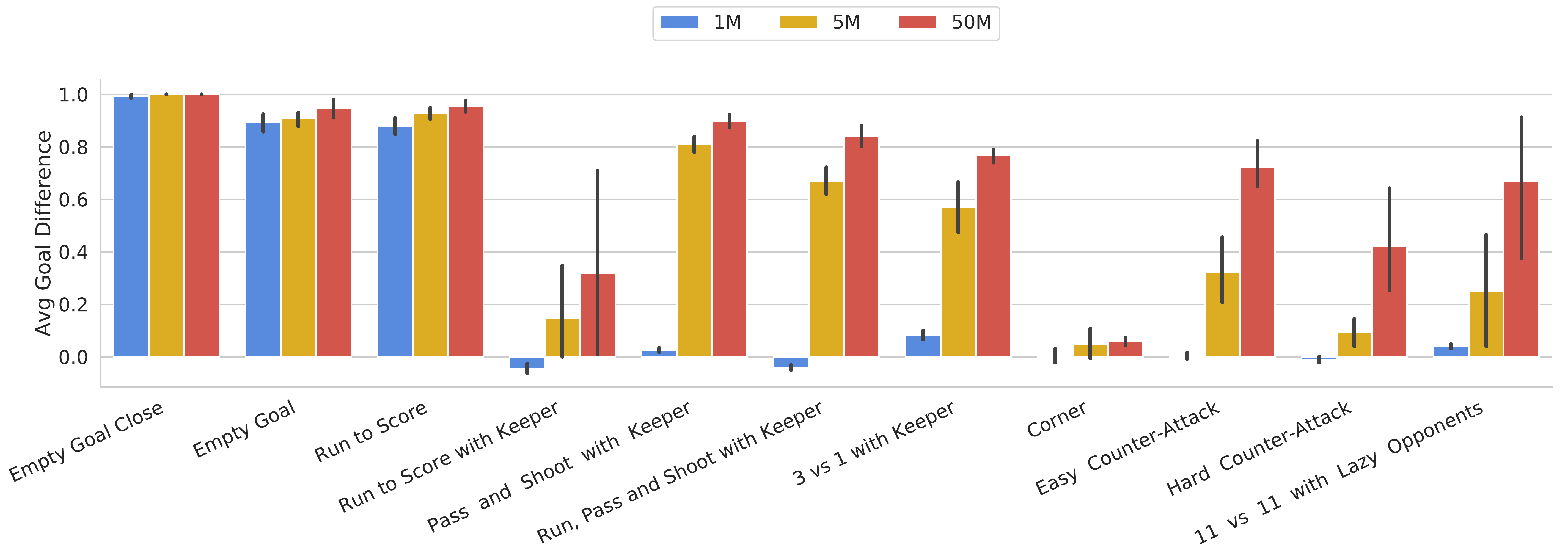}
    \caption{Average Goal Difference on  \academy for PPO with \textsc{Scoring} reward.}
    \label{fig:academy_ppo_scoring}    
\end{figure*}
\section{Football Academy}
\looseness=-1Training agents for the \benchmarks can be challenging.
To allow researchers to quickly iterate on new research ideas, we also provide the \academy: a diverse set of scenarios of varying difficulty.
These 11 scenarios (see Figure~\ref{fig:academy_scenarios} for a selection) include several variations where a single player has to score against an empty goal (\emph{Empty Goal Close}, \emph{Empty Goal}, \emph{Run to Score}), a number of setups where the controlled team has to break a specific defensive line formation (\emph{Run to Score with Keeper}, \emph{Pass and Shoot with Keeper}, \emph{3 vs 1 with Keeper}, \emph{Run, Pass and Shoot with Keeper}) as well as some standard situations commonly found in football games (\emph{Corner}, \emph{Easy Counter-Attack}, \emph{Hard Counter-Attack}).
For a detailed description, we refer to the Appendix.
Using a simple API, researchers can also easily define their own scenarios and train agents to solve them.

\subsection{Experimental Results}
Based on the same experimental setup as for the \benchmarks, we provide experimental results for both PPO and IMPALA for the \academy scenarios in Figures~\ref{fig:academy_impala_scoring}, \ref{fig:academy_ppo_scoring}, \ref{fig:academy_impala_checkpoint}, and \ref{fig:academy_ppo_checkpoint} (the last two are provided in the Appendix).
We note that the maximum average scoring performance is 1 (as episodes end in the \academy scenarios after scoring) and that scores may be negative as agents may score own goals and as the opposing team can score in the \emph{Corner} scenario.

The experimental results indicate that the \academy provides a set of diverse scenarios of different difficulties suitable for different computational constraints.
The scenarios where agents have to score against the empty goal (\emph{Empty Goal Close}, \emph{Empty Goal}, \emph{Run to Score}) appear to be very easy and can be solved both PPO and IMPALA with both reward functions using only 1M steps.
As such, these scenarios can be considered ``unit tests'' for reinforcement learning algorithms where one can obtain reasonable results within minutes or hours instead of days or even weeks.
The remainder of the tasks includes scenarios for which both PPO and IMPALA appear to require between 5M to 50M steps for progress to occur (with minor differences between the \textsc{Scoring} and \textsc{Checkpoint}) rewards).
These harder tasks may be used to quickly iterate on new research ideas on single machines before applying them to the \benchmarks (as experiments should finish within hours or days).
Finally, the \textsc{Corner} appears to be the hardest scenario (presumably as one has to face a full squad and the opponent is also allowed to score). 

\section{Promising Research Directions}
In this section we briefly discuss a few initial experiments related to three research topics which have recently become quite active in the reinforcement learning community: self-play training, multi-agent learning, and representation learning for downstream tasks.
This highlights the research potential and flexibility of the Football Environment.

\subsection{Multiplayer Experiments}
The Football Environment provides a way to train against different opponents, such as built-in AI or other trained agents.
Note this allows, for instance, for self-play schemes.
When a policy is trained against a fixed opponent, it may exploit its particular weaknesses and, thus, it may not generalize well to other adversaries.
We conducted an experiment to showcase this in which a first model $A$ was trained against a built-in AI agent on the standard 11 vs 11 medium scenario.
Then, another agent $B$ was trained against a frozen version of agent $A$ on the same scenario. While $B$ managed to beat $A$ consistently, its performance against built-in AI was poor.
The numerical results showing this lack of transitivity across the agents are presented in Table~\ref{tab:transitivity}.

\begin{table}[h]
\caption{Average goal difference $\pm$ one standard deviation across 5 repetitions of the experiment.}
\centering
\label{tab:transitivity}
\begin{tabular}{lr}
\toprule
$A$ vs built-in AI & $4.25 \pm 1.72$ \\
$B$ vs $A$ & $11.93 \pm 2.19$ \\
$B$ vs built-in AI & $-0.27 \pm 0.33$ \\
\bottomrule
\end{tabular}

\end{table}

\subsection{Multi-Agent Experiments}
The environment also allows for controlling several players from one team simultaneously, as in multi-agent reinforcement learning.
We conducted experiments in this setup with the \emph{3 versus 1 with Keeper} scenario from Football Academy.
We varied the number of players that the policy controls from 1 to 3, and trained with Impala. 

As expected, training is initially slower when we control more players, but the policies seem to eventually learn more complex behaviors and achieve higher scores.
Numerical results are presented in Table~\ref{tab:multiagent}.

\begin{table}
\caption{Scores achieved by the policy controlling 1, 2 or 3 players respectively, after 5M and 50M steps of training.}
\centering
\begin{tabular}{lrr}
\toprule
\textbf{Players controlled} & \textbf{5M steps} & \textbf{50M steps} \\ \midrule
1 & $0.38 \pm 0.23$ & $0.68 \pm 0.03$ \\
2 & $0.17 \pm 0.18$ & $0.81 \pm 0.17$ \\
3 & $0.26 \pm 0.11$ & $0.86 \pm 0.08$ \\
\bottomrule
\end{tabular}
\label{tab:multiagent}
\end{table}

\subsection{Representation Experiments}

Training the agent directly from raw observations, such as pixels, is an exciting research direction. 
While it was successfully done for Atari, it is still an open challenge for most of the more complex and realistic environments. 
In this experiment, we compare several representations available in the \engine. 
\emph{Pixels gray} denotes the raw pixels from the game, which are resized to $72 \times 96$ resolution and converted to grayscale.
While pixel representation takes significantly longer time to train, as shown in Table~\ref{tab:representation}, learning eventually takes place (and it actually outperforms hand-picked extensive representations like `Floats').
The results were obtained using Impala with Checkpoint reward on the easy 11 vs.\ 11 benchmark. 

\begin{table}
\centering
\caption{Average goal advantages per representation.}
\label{tab:representation}
\begin{tabular}{lrr}
\toprule
\textbf{Representation} & \textbf{100M steps} & \textbf{500M steps} \\ \midrule
Floats &  $2.42 \pm 0.46$ & $5.73 \pm 0.28$ \\
Pixels gray & $-0.82 \pm 0.26$ &  $7.18 \pm 0.85$ \\
SMM & $5.94 \pm 0.86$ & $9.75 \pm 4.15$ \\
SMM stacked & $7.62 \pm 0.67$ & $12.89 \pm 0.51$ \\

\bottomrule
\end{tabular}
\end{table}

\section{Conclusions}
\looseness=-1In this paper, we presented the \emph{Google Research Football Environment}, a novel open-source reinforcement learning environment for the game of football.
It is challenging and accessible, easy to customize, and it has specific functionality geared towards research in reinforcement learning.
We provided the \engine, a highly optimized C++ football simulator, the \benchmarks, a set of reference tasks to compare different reinforcement learning algorithms, and the \academy, a set of progressively harder scenarios.
We expect that these components will be useful for investigating current scientific challenges like self-play, sample-efficient RL, sparse rewards, and model-based RL.

\clearpage
\section*{Acknowledgement}
We wish to thank Lucas Beyer, Nal Kalchbrenner, Tim Salimans and the rest of the Google Brain team for helpful discussions, comments, technical help and code contributions. We would also like to thank Bastiaan Konings Schuiling, who authored and open-sourced the original version of this game.

\bibliographystyle{aaai}
{\small \bibliography{main}}

\clearpage

\appendix

\section{Hyperparameters \& Architectures}
\label{app:hparams}

\begin{figure}
    \includegraphics[width=\columnwidth]{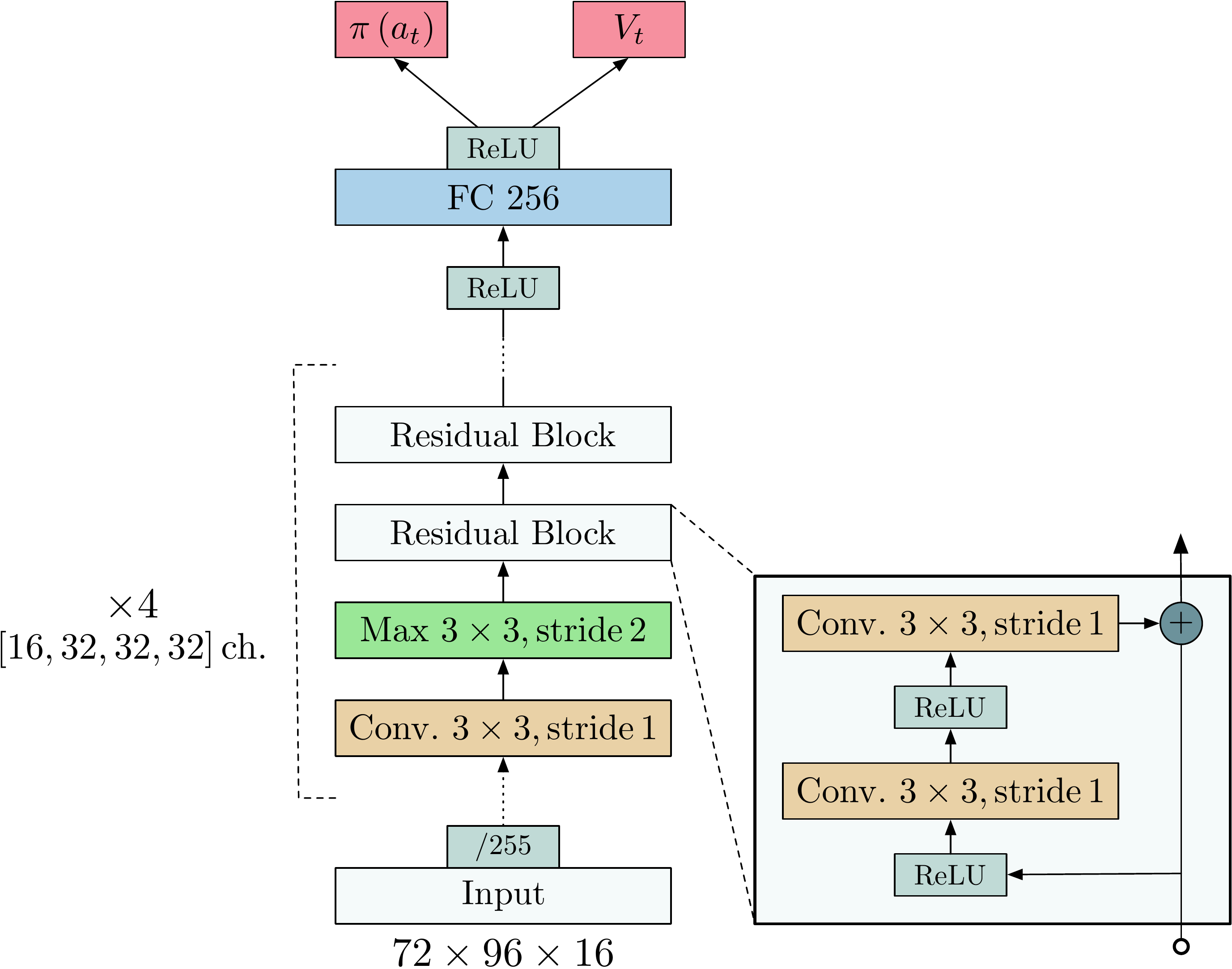}
    \caption{Architecture used for IMPALA and PPO experiments. For Ape-X DQN, a similar network is used but the outputs are Q-values.}
    \label{fig:impala_architecture}
\end{figure}

For our experiments, we used three algorithms (IMPALA, PPO, Ape-X DQN) that are described below. The model architecture we use is inspired by Large architecture from~\citep{espeholt2018impala} and is depicted in Figure~\ref{fig:impala_architecture}. 
Based on the "Representation Experiments", we selected the stacked Super Mini Map\ref{sec:smm} as the default representation used in all \benchmarks and \academy experiments.
In addition we have three other representations.

For each of the six considered settings (three \benchmarks and two reward functions), we run five random seeds for $500$ million steps each. For \academy, we run five random seeds in all $11$ scenarios for $50$ million steps.

\paragraph{Hyperparameter search}
For each of IMPALA, PPO and Ape-X DQN, we performed two hyperparameter searches: one for \textsc{Scoring} reward and one for \textsc{Checkpoint} reward. For the search, we trained on easy difficulty. Each of 100 parameter sets was repeated with 3 random seeds. For each algorithm and reward type, the best parameter set was decided based on average performance -- for IMPALA and Ape-X DQN after 500M, for PPO after 50M. After the search, each of the best parameter sets was used to run experiments with 5 different random seeds on all scenarios. Ranges that we used for the procedure can be found in Table~\ref{tab:impala_hparams_values} for IMPALA, Table~\ref{tab:ppo_hparams_values} for PPO and Table~\ref{tab:dqn_hparams_values} for DQN.

\paragraph{IMPALA}
Importance Weighted Actor-Learner Architecture \citep{espeholt2018impala} is a highly scalable algorithm that decouples acting from learning. Individual workers communicate trajectories of experience to the central learner, instead of sending gradients with respect to the current policy. In order to deal with off-policy data, IMPALA introduces an actor-critic update for the learner called V-trace. Hyper-parameters for IMPALA are presented in Table~\ref{tab:impala_hparams_values}.

\paragraph{PPO}
Proximal Policy Optimization \citep{schulman2017proximal} is an online policy gradient algorithm which optimizes the clipped surrogate objective.
In our experiments we use the implementation from the OpenAI Baselines~\citep{baselines}, and run it over 16 parallel workers.
Hyper-parameters for PPO are presented in Table~\ref{tab:ppo_hparams_values}.

\paragraph{Ape-X DQN}
Q-learning algorithms are popular among reinforcement learning researchers.
Accordingly, we include a member of the DQN family in our comparison. In particular, we chose Ape-X DQN~\citep{Horgan2018DistributedPE}, a highly scalable version of DQN.
Like IMPALA, Ape-X DQN decouples acting from learning but, contrary to IMPALA, it uses a distributed replay buffer and a variant of Q-learning consisting of dueling network architectures~\citep{pmlr-v48-wangf16} and double Q-learning~\citep{van2016deep}.

Several hyper-parameters were aligned with IMPALA. These includes unroll length and $n$-step return, the number of actors and the discount factor $\gamma$. For details, please refer to the Table~\ref{tab:dqn_hparams_values}.

\section{Numerical Results for the \benchmarks}
In this section we provide for comparison the means and std values of $5$ runs for all algorithms in \benchmarks. Table~\ref{tab:benchmark_scoring} contains the results for the runs with \textsc{Scoring} reward while Table~\ref{tab:benchmark_checkpoint} contains the results for the runs with \textsc{Checkpoint} reward.
Those numbers were presented in the main paper in Figure~\ref{fig:challenges_both_rewards}.

\begin{table}[h]
\caption{Benchmark results for \textsc{Scoring} reward.}
\resizebox{\columnwidth}{!}{

\begin{tabular}{lrrr}
\toprule
\textsc{Model} &     \textsc{Easy} &   \textsc{Medium} &     \textsc{Hard} \\
\midrule
      PPO @20M &   $0.05 \pm 0.13$ &  $-0.74 \pm 0.08$ &  $-1.32 \pm 0.12$ \\
      PPO @50M &   $0.09 \pm 0.13$ &  $-0.84 \pm 0.10$ &  $-1.39 \pm 0.22$ \\
   IMPALA @20M &  $-0.01 \pm 0.10$ &  $-0.89 \pm 0.34$ &  $-1.38 \pm 0.22$ \\
  IMPALA @500M &   $5.14 \pm 2.88$ &  $-0.36 \pm 0.11$ &  $-0.47 \pm 0.48$ \\
      DQN @20M &  $-1.17 \pm 0.31$ &  $-1.63 \pm 0.11$ &  $-2.12 \pm 0.33$ \\
     DQN @500M &   $8.16 \pm 1.05$ &   $2.01 \pm 0.27$ &   $0.27 \pm 0.56$ \\
\bottomrule
\end{tabular}

}
\label{tab:benchmark_scoring}
\end{table}

\begin{table}[h]
\caption{Benchmark results for \textsc{Checkpoint} reward.}
\resizebox{\columnwidth}{!}{
\begin{tabular}{lrrr}
\toprule
\textsc{Model} &     \textsc{Easy} &   \textsc{Medium} &     \textsc{Hard} \\
\midrule
      PPO @20M &   $6.23 \pm 1.25$ &   $0.38 \pm 0.49$ &  $-1.25 \pm 0.09$ \\
      PPO @50M &   $8.71 \pm 0.72$ &   $1.11 \pm 0.45$ &  $-0.75 \pm 0.13$ \\
   IMPALA @20M &  $-1.00 \pm 0.34$ &  $-1.86 \pm 0.13$ &  $-2.24 \pm 0.08$ \\
  IMPALA @500M &  $12.83 \pm 1.30$ &   $5.54 \pm 0.90$ &   $3.15 \pm 0.37$ \\
      DQN @20M &  $-1.15 \pm 0.37$ &  $-2.04 \pm 0.45$ &  $-2.22 \pm 0.19$ \\
     DQN @500M &   $7.06 \pm 0.85$ &   $2.18 \pm 0.25$ &   $1.20 \pm 0.40$ \\
\bottomrule
\end{tabular}

}
\label{tab:benchmark_checkpoint}
\end{table}

\clearpage

\begin{table*}
\caption{IMPALA: ranges used during the hyper-parameter search and the final values used for experiments with scoring and checkpoint rewards.}
\begin{center}
\begin{tabular}{lrrr}
\toprule
\textbf{Parameter}  &\textbf{Range} & \textbf{Best - Scoring} & \textbf{Best - Checkpoint} \\ \midrule
Action Repetitions & 1 & 1 & 1 \\
Batch size & 128 & 128 & 128 \\
Discount Factor ($\gamma$) & $\{.99, .993, .997, .999\}$ & .993 & .993 \\
Entropy Coefficient & Log-uniform $(1\mathrm{e}{-6}$, $1\mathrm{e}{-3})$ & 0.00000521 & 0.00087453 \\
Learning Rate & Log-uniform $(1\mathrm{e}{-5}$, $1\mathrm{e}{-3})$ & 0.00013730 & 0.00019896 \\
Number of Actors & 500 & 500 & 500 \\
Optimizer & Adam & Adam & Adam \\
Unroll Length/$n$-step & $\{16, 32, 64\}$ & 32 & 32 \\
Value Function Coefficient &.5 & .5 & .5 \\
\bottomrule
\end{tabular}

\label{tab:impala_hparams_values}
\end{center}
\end{table*}

\begin{table*}[h]
\caption{PPO: ranges used during the hyper-parameter search and the final values used for experiments with scoring and checkpoint rewards.}
\begin{center}

\begin{tabular}{lrrr}
\toprule
\textbf{Parameter} & \textbf{Range} & \textbf{Best - Scoring} & \textbf{Best - Checkpoint} \\ \midrule
Action Repetitions & 1 & 1 & 1 \\
Clipping Range  & Log-uniform $(.01, 1)$ & .115 & .08 \\
Discount Factor ($\gamma$) & $\{.99, .993, .997, .999\}$ & .997 & .993 \\
Entropy Coefficient & Log-uniform $(.001, .1)$ & .00155 & .003 \\
GAE ($\lambda$) & .95 & .95 & .95 \\
Gradient Norm Clipping & Log-uniform $(.2, 2)$ & .76 & .64 \\
Learning Rate  & Log-uniform $(.000025, .0025)$  & .00011879 & .000343 \\
Number of Actors & 16 & 16 & 16 \\
Optimizer & Adam & Adam & Adam \\
Training Epochs per Update & $\{2, 4, 8\}$ & 2 & 2 \\
Training Mini-batches per Update & $\{2, 4, 8\}$ & 4 & 8 \\
Unroll Length/$n$-step & $\{16, 32, 64, 128, 256, 512\}$ & 512 & 512 \\
Value Function Coefficient & .5 & .5 & .5 \\
\bottomrule
\end{tabular}
\label{tab:ppo_hparams_values}
\end{center}
\end{table*}

\begin{table*}
\caption{DQN: ranges used during the hyper-parameter search and the final values used for experiments with scoring and checkpoint rewards.}

\begin{center}
\begin{tabular}{lrrr}
\toprule
\textbf{Parameter} & \textbf{Range} & \textbf{Best - Scoring} & \textbf{Best - Checkpoint} \\ \midrule
Action Repetitions & 1 & 1 & 1 \\
Batch Size & 512  & 512 & 512 \\
Discount Factor ($\gamma$) & $\{.99, .993, .997, .999\}$ & .999 & .999 \\
Evaluation $\epsilon$ & .01 & .01 & .01 \\
Importance Sampling Exponent & $\{0., .4, .5, .6, .8, 1.\}$ & 1. & 1.  \\
Learning Rate & Log-uniform $(1\mathrm{e}{-7}$, $1\mathrm{e}{-3})$ & .00001475 & .0000115 \\
Number of Actors & 150 & 150 & 150 \\
Optimizer & Adam & Adam & Adam \\
Replay Priority Exponent & $\{0., .4, .5, .6, .7, .8\}$  & .0 & .8 \\
Target Network Update Period & 2500 & 2500 & 2500 \\
Unroll Length/$n$-step & $\{16, 32, 64, 128, 256, 512\}$ & 16 & 16 \\
\bottomrule
\end{tabular}
\label{tab:dqn_hparams_values}
\end{center}
\end{table*}

\begin{figure*}[h]
    \centering
    \includegraphics[width=\textwidth]{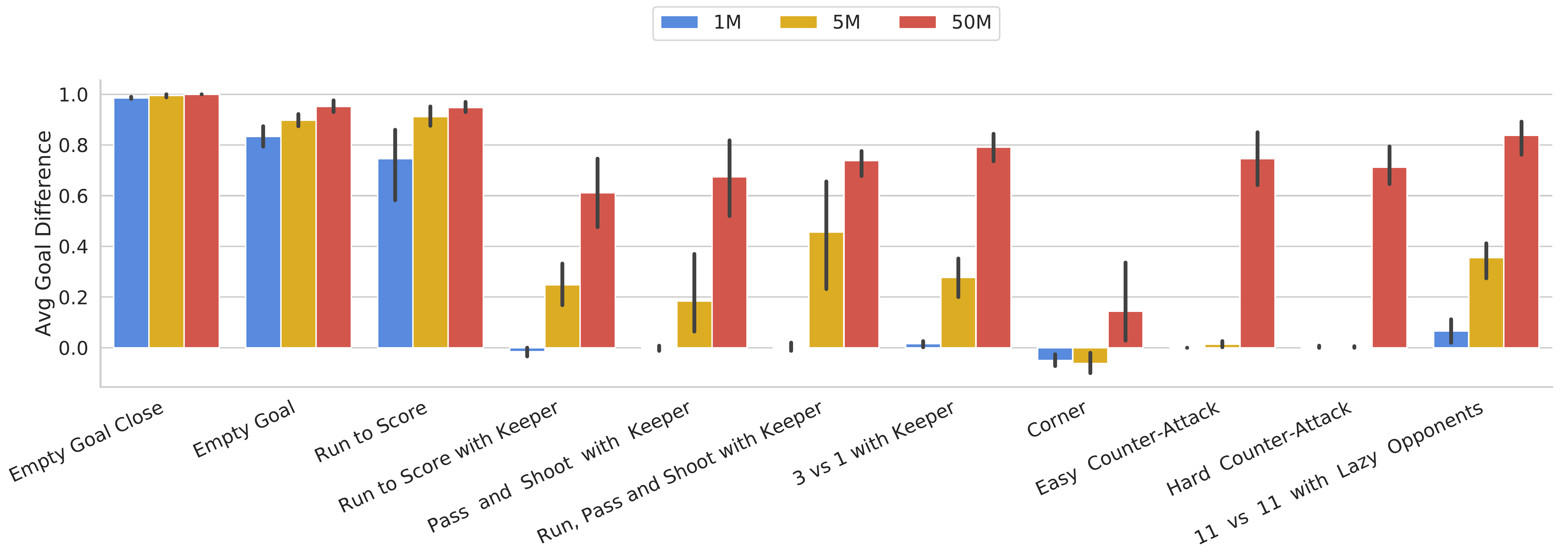}
    
    \caption{Average Goal Difference on  \academy for IMPALA with \textsc{Checkpoint} reward.}
    \label{fig:academy_impala_checkpoint}    
    
\end{figure*}

\begin{figure*}[h]
    \centering
    \includegraphics[width=\textwidth]{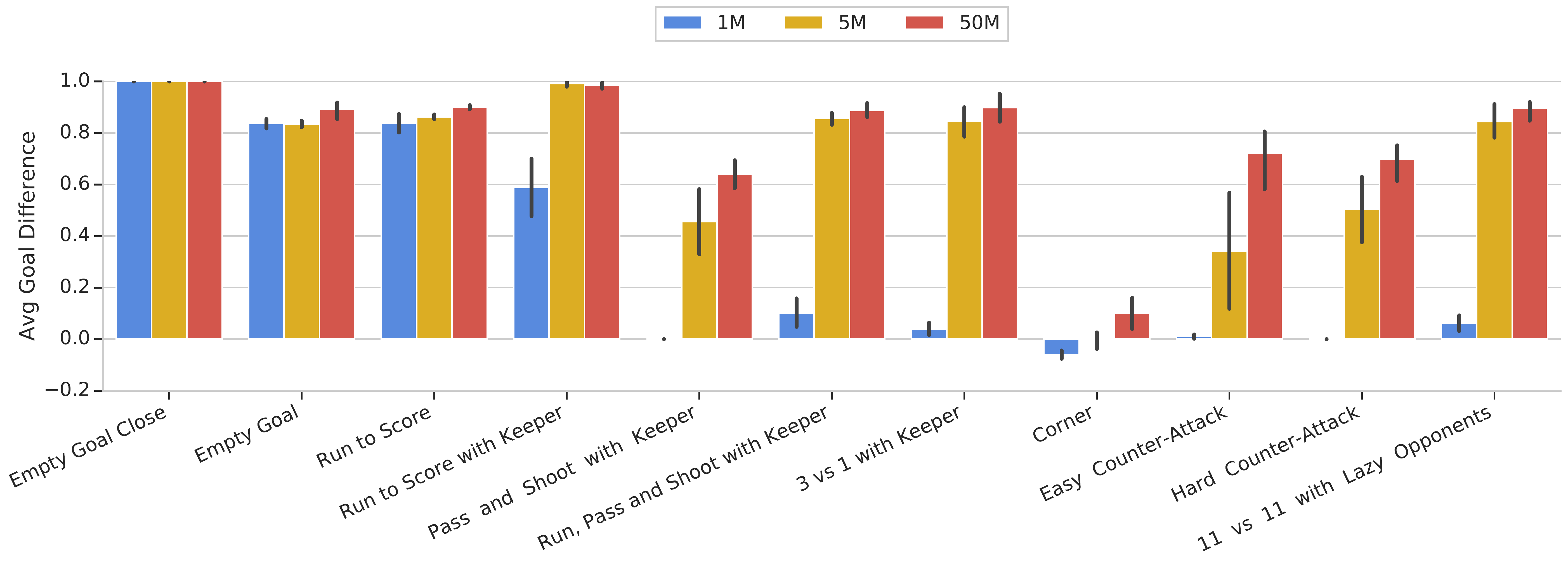}
    \caption{Average Goal Difference on  \academy for PPO with \textsc{Checkpoint} reward. Scores for v$1.x$ (All other results in this paper are for v$2.x$, but for this plot the experiment didn't finish. Please check arxiv for the full v$2.x$ results)}
    \label{fig:academy_ppo_checkpoint}
\end{figure*}

\begin{table*}[h!]
\caption{Description of the default \academy scenarios. If not specified otherwise, all scenarios end after 400 frames or if the ball is lost, if a team scores, or if the game is stopped (\emph{e.g.} if the ball leaves the pitch or if there is a free kick awarded).The difficulty level is 0.6 (\emph{i.e.}, medium).}
\renewcommand*{\arraystretch}{1.5}
\begin{center}

\begin{tabular}{p{5cm}p{9cm}}
\toprule
\textbf{Name} & \textbf{Description} \\ \midrule
\textit{Empty Goal Close} & Our player starts inside the box with the ball, and needs to score against an empty goal. \\
\textit{Empty Goal} & Our player starts in the middle of the field with the ball, and needs to score against an empty goal. \\
\textit{Run to Score} & Our player starts in the middle of the field with the ball, and needs to score against an empty goal. Five opponent players chase ours from behind. \\
\textit{Run to Score with Keeper} & Our player starts in the middle of the field with the ball, and needs to score against a keeper. Five opponent players chase ours from behind. \\
\textit{Pass and Shoot with Keeper} & Two of our players try to score from the edge of the box, one is on the side with the ball, and next to a defender. The other is at the center, unmarked, and facing the opponent keeper. \\
\textit{Run, Pass and Shoot with Keeper} & Two of our players try to score from the edge of the box, one is on the side with the ball, and unmarked. The other is at the center, next to a defender, and facing the opponent keeper. \\
\textit{3 versus 1 with Keeper} & Three of our players try to score from the edge of the box, one on each side, and the other at the center. Initially, the player at the center has the ball, and is facing the defender. There is an opponent keeper. \\
\textit{Corner} & Standard corner-kick situation, except that the corner taker can run with the ball from the corner. The episode does not end if possession is lost.\\
\textit{Easy Counter-Attack} & 4 versus 1 counter-attack with keeper; all the remaining players of both teams run back towards the ball. \\
\textit{Hard Counter-Attack} & 4 versus 2 counter-attack with keeper; all the remaining players of both teams run back towards the ball. \\
\textit{11 versus 11 with Lazy Opponents} & Full 11 versus 11 game, where the opponents cannot move but they can only intercept the ball if it is close enough to them. Our center-back defender has the ball at first.  The maximum duration of the episode is 3000 frames instead of 400 frames.\\
\bottomrule
\end{tabular}
\label{tab:scenario_description}
\end{center}
\end{table*}

\end{document}